\documentclass{article}

\usepackage[final]{tackling_climate_workshop_style}

\usepackage[utf8]{inputenc} 
\usepackage[T1]{fontenc}    
\usepackage{hyperref}       
\usepackage{url}            
\usepackage{booktabs}       
\usepackage{amsfonts}       
\usepackage{nicefrac}       
\usepackage{microtype}      
\usepackage{graphicx}
\usepackage{multirow}
\usepackage{tabularray}
\usepackage{subcaption}
\usepackage{natbib}

\title{Improving Power Plant CO$_2$ Emission Estimation with Deep Learning and Satellite/Simulated Data}

\author{%
  Dibyabha Deb\\
  Manipal Institute of Technology, India\\
  \texttt{dibyabha.deb@learner.manipal.edu} \\
  \And
  Kamal Das \\
  IBM Research, India \\
  \texttt{kdas3@in.ibm.com} \\
}

\begin{document}

\maketitle

\begin{abstract}
CO$_2$ emissions from power plants, as significant super emitters, contribute substantially to global warming. Accurate quantification of these emissions is crucial for effective climate mitigation strategies. While satellite-based plume inversion offers a promising approach, challenges arise from data limitations and the complexity of atmospheric conditions. This study addresses these challenges by (a) expanding the available dataset through the integration of NO$_2$ data from Sentinel-5P, generating continuous XCO$_2$ maps, and incorporating real satellite observations from OCO-2/3 for over 71 power plants in data-scarce regions; and (b) employing a customized U-Net model capable of handling diverse spatio-temporal resolutions for emission rate estimation. Our results demonstrate significant improvements in emission rate accuracy compared to previous methods \cite{le2024deep}. By leveraging this enhanced approach, we can enable near real-time, precise quantification of major CO$_2$ emission sources, supporting environmental protection initiatives and informing regulatory frameworks. 

\end{abstract}
\vspace{-10pt}
\section{Introduction}
\vspace{-10pt}
    Over the past two decades, greenhouse gases (GHG) emissions from burning fossil fuels by electricity industry has increased its carbon dioxide (CO$_2$) emissions by 53\% worldwide \cite{newell2021global}. Experts expect that more than half of all future carbon emissions will come from this sector \cite{tong2019committed}. Understanding power plant emissions ("super emitters"), especially in regions where proper accounting mechanisms are lacking, is crucial for identifying sources and quantifying emissions to support carbon-neutrality pathways. Estimation of carbon emissions to the atmosphere has generally been performed using two complementary approaches: “bottom-up” and “top-down” methods \cite{elguindi2020intercomparison}. Bottom-up methods aggregate source-specific CO$_2$ flux estimates to form a total emission inventory based on activity data and emission models \cite{gurney2012quantification}. These inventories can be highly resolved in both space and time, but they are prone to systematic errors, and their uncertainties are not well known \cite{andres2014new} and their time lags for more than years with respect to real time is another concern. Top-down methods infer quantitative information on surface CO$_2$ fluxes from variations in atmospheric CO$_2$ concentration observations through inverse modeling with atmospheric tracer transport models \cite{wu2018joint}. Recently, there has been an increase in atmospheric observations through satellite observations \cite{boesch2021monitoring}, driven by the growing number of dedicated satellites, advancements in sensor technology, and improved spatial resolution. This has heightened interest in using atmospheric observations to derive CO$_2$ emissions. Traditional inverse modeling systems, often relying on Lagrangian Particle Dispersion Models (LPDMs), are a popular approach to quantify CO$_2$ emission rates using atmospheric observations \cite{fillola2023machine}. However, increased volume of atmospheric observations strained the computational capacity of these traditional methods along with challenges remain in the spatial allocation of emissions due to uncertainties in prior fluxes and atmospheric transport models. 
    
    In recent years, efforts have been made to use data-driven learning approaches to detect plumes and their shapes, and then invert CO$_2$ plumes from major emission sources. However, many studies rely on simulated data to learn the complex concentration-to-flux conversion, and there are very few studies exploring similar techniques using observed data from satellites \cite{le2024deep}. To enhance the current data-driven understanding, three main challenges must be addressed: (1) Observed data from satellites, such as carbon dioxide concentration (XCO$_2$) measurements, are often limited, with only 10\% of the data being clean and temporally very sparse \cite{nassar2023intelligent}. (2) The robustness and generalizability of data-driven approaches are limited when only simulated data are used, due to inconsistencies in emission distributions and the lack of data representing diverse power plant profiles. Additionally, CO$_2$ plumes are notoriously difficult to invert due to various factors such as missing data and signal-to-noise ratio (SNR) issues. (3) Super emitters, like power plants, do not report emissions in a timely manner or at a fine temporal frequency, which poses a labeling challenge for supervised learning models aimed at learning emission rates at finer temporal resolutions. In this work, we aim to address (a) the sparsity of satellite observation data by employing a machine learning approach, (b) create a novel dataset by merging simulated data with satellite observations, and finally (c) customize a U-Net model for learning CO$_2$ concentration-to-flux conversion for power plant emission rates. 

\vspace{-10pt}
\section{Model to Estimate Emission Rate}
\vspace{-10pt}
We have employed two distinct deep learning architectures to estimate CO$_2$ emission rates for power plants: the Convolutional Neural Network (CNN) and the U-Net regression model. Each architecture is tailored to capture spatial patterns and relationships within the input data, aiming to enhance the accuracy of emissions predictions. We briefly summarize these models below:

\textbf{Baseline Model - CNN:} The CNN model as adapted from \cite{le2024deep} uses a multi-layered architecture consisting of convolution layers that detect features, max-pooling for dimensionality reduction and retaining key features, batch normalization for training stability and dropout to prevent over-fitting. The final dense layer with Leaky ReLU activation produces the predictions. This model serves as the baseline for our performance evaluation.

\begin{figure}[h!]
\vspace{-8pt}
    \centering
    \begin{subfigure}[b]{0.49\textwidth}
        \centering
        \includegraphics[width=\textwidth]{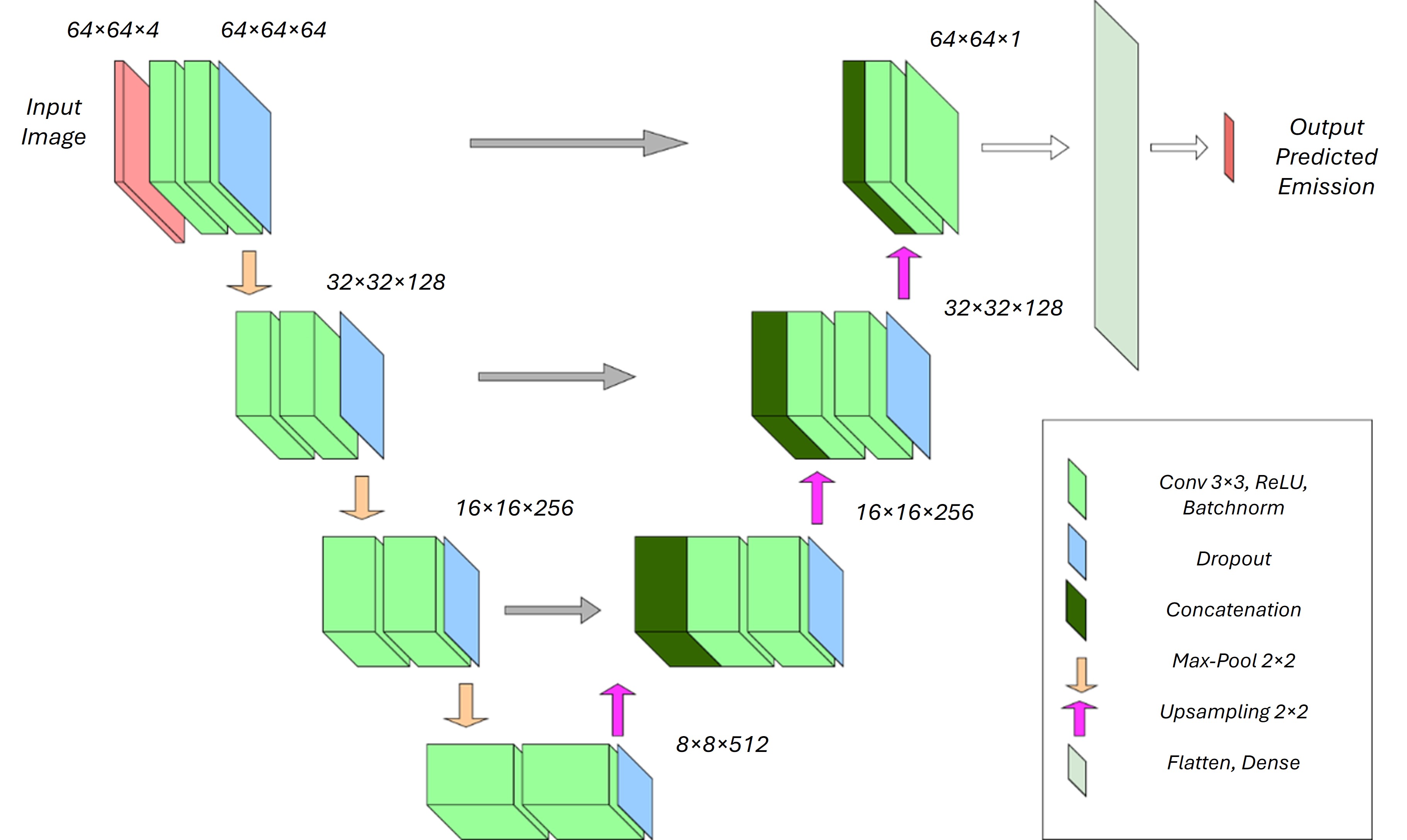}
        \caption{U-Net model architecture}
        \label{fig:first}
    \end{subfigure}
    \hfill
    \begin{subfigure}[b]{0.49\textwidth}
        \centering
        \includegraphics[width=\textwidth]{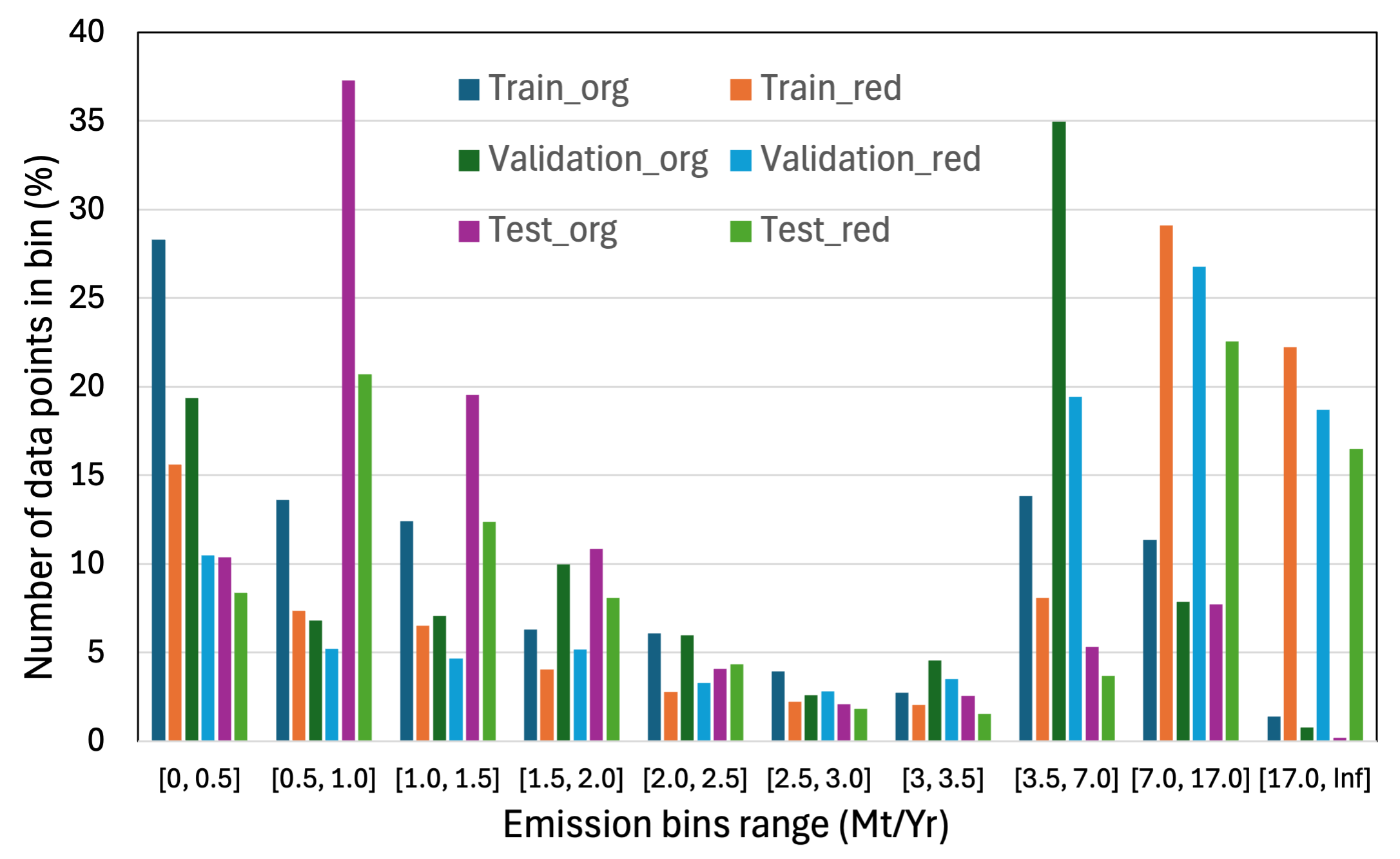}
        \caption{Dataset distribution}
        \label{fig:second}
    \end{subfigure}
    \caption{(a) The Model Architecture of the U-Net Regression model for CO2 concentration to emission estimation; (b) Annual emission rates by bins for the combined dataset with original and redistributed
splits across training, validation, and test sets.}
    \label{fig:main}
\vspace{-8pt}
\end{figure}

\textbf{Proposed U-Net Regression Model:}
The U-Net architecture includes an encoder-decoder structure that efficiently handles feature extraction and spatial context preservation. The model used for the regression task has been shown in the Figure \ref{fig:first} where the encoder captures contextual information progressively through a series of convolution, pooling layers, batch normalization and dropout (0.2) layers to train the model \cite{ronneberger2015u}. The output of the decoder layer is taken as input into a final dense layer to predict the emission rate. To assess the performance of the model, we used several metrics: Absolute Error, Absolute Relative Error, Mean Absolute Error (MAE), Root Mean Squared Error (RMSE), and R$^2$ values. Each metric helped evaluate how well the model performed on the same test dataset.


\vspace{-10pt}
\section{Power Plant Emission: Data and Results}
\vspace{-10pt}
\textbf{Proposed Satellite observations data: } Satellite observations XCO$_2$, derived from the combined data of OCO2/3 satellites and subject to rigorous quality control, often exhibit sparse spatial and temporal distribution. To facilitate effective analytics, continuous, gridded XCO$_2$ maps with regular temporal frequency are essential. Therefore, we have implemented a machine learning approach, analogous to the method presented in \cite{das2023machine}, to generate daily XCO$_2$ maps with a spatial resolution of 1km$\times$1km across Kingdom of Saudi Arabia (KSA). The nitrogen dioxide (NO$_2$) data utilized in this study were sourced from Sentinel-5P/TROPOMI (TROPOspheric Monitoring Instrument) Level 3 Offline (OFFL) satellite imagery (COPERNICUS/S5P/OFFL/L3\_NO2), offering a spatial resolution of 3.5km$\times$5.5km and daily temporal frequency. However, due to cloud cover and other factors, the data contained missing values. To generate regular gridded maps without missing data, we implemented a series of post-processing steps designed to fill missing values and enhance spatial resolution to 1km$\times$1km using appropriate downscaling approach \cite{kim2021importance}. High-resolution meteorological data was obtained from TWC sources \cite{bickel2008verification}, providing hourly weather variables at a spatial resolution of 4km$\times$4km. To align with the desired spatial resolution of 1km$\times$1km, we downscaled wind velocities using interpolation techniques. For the KSA region, KAUST University published a dataset based on a bottom-up approach, reporting the annual emissions for 71 power plants in 2020, along with their geolocations \cite{hamieh2022quantification}. To create power plant-specific datasets for model training, validation, and testing, we used the power plant geolocations as the centroid pixel and created 64 $\times$ 64 patch size images with XCO$_2$, NO$_2$, and both wind velocity components at a daily frequency. CO$_2$ emissions are represented as a scalar value for each power plant, indicating the annual emission rate. We have disaggregated this annual rate to a daily level using a truncated version of the approach outlined in \cite{crippa2020high}. To achieve this, we used daily trace gas measurements, energy production data, and weather profiles. Daily feature images and target emission rates for each power plant were merged to create satellite observation data for the KSA region. 

\textbf{Simulated data:} The existing dataset was taken from the previous paper \cite{le2024deep} which comprised of simulated satellite imagery and atmospheric variables, including column average of carbon dioxide concentration (XCO$_2$), wind speed fields (u and v compoents), and nitrogen dioxide concentration (NO$_2$). These measurements span multiple power plant regions from the eastern part of Germany and nearby regions, facilitating a comprehensive analysis of emissions data (more detail in Appendix).

\textbf{Combined Simulated and Satellite observations data:} The satellite-derived data have a spatial resolution of 1km$\times$1km with daily temporal frequency, while the simulated data from the previous paper \cite{le2024deep} have a spatial resolution of 2km$\times$2km and include emission rates at finer temporal frequencies. We merged these datasets into a single dataset and redistributed it into training, validation, and test splits based on emission levels. Figure \ref{fig:second} shows the comparison between the original simulated data distribution (train\_org, valid\_org, test\_org) and the novel combined dataset distribution (train\_red, valid\_red, test\_red) on train, validation and test split respectively. The significant differences in each bin will help the model address overfitting and improve performance on unseen data. The combined dataset’s differences in spatial and temporal resolutions are handled by both models. 



\textbf{Training and calibration: } The loss functions used during our model training include MAE, Mean Absolute Percentage Error (MAPE), Mean Squared Error (MSE), and Huber Loss. Prior work \cite{le2024deep} employed MAE and MAPE in their CNN model. In contrast, we utilized all four loss functions and averaged the predictions from each. This comprehensive approach leverages the strengths of each loss function, offering a more nuanced evaluation of model performance. Previous studies \cite{le2024deep} have suggested that segmented plumes may not significantly enhance CNN performance. Based on these findings, we did not incorporate plume-segmented maps in our analysis. Our study utilized the same dataset framework as prior research, with feature data consisting of XCO$_2$, NO$_2$, wind vectors (U and V), and the emission rate as the target variable. 

\begin{table}
\centering
\tiny
\caption{Model performance assessment summary with the best-performing model metrics highlighted in bold for both CNN and U-Net models across three dataset scenarios}
\begin{tblr}{
  width = \linewidth,
  colspec = {Q[100]Q[70]Q[58]Q[69]Q[58]Q[63]Q[69]Q[69]Q[96]Q[104]Q[90]Q[58]Q[35]},
  cells = {c},
  cell{1}{1} = {r=2}{},
  cell{1}{2} = {r=2}{},
  cell{1}{3} = {c=3}{0.185\linewidth},
  cell{1}{6} = {c=3}{0.201\linewidth},
  cell{1}{9} = {r=2}{},
  cell{1}{10} = {r=2}{},
  cell{1}{11} = {r=2}{},
  cell{1}{12} = {r=2}{},
  cell{1}{13} = {r=2}{},
  cell{3}{1} = {r=2}{},
  cell{5}{1} = {r=2}{},
  cell{7}{1} = {r=2}{},
  vlines,
  hline{1,3,5,7,9} = {-}{},
  hline{2} = {3-8}{},
  hline{4,6,8} = {2-13}{},
}
Dataset               & Model & Absolute
  Error (Mt/Yr) &        &      & Absolute
  Relative Error (\%) &        &        & {Median Absolute \\Error (Mt/Yr)} & {Median Absolute \\Relative
  Error (\%)} & {Mean Absolute \\Error (Mt/Yr)} & {RMSE \\(Mt/Yr)} & $R^2$ \\
                      &       & 25\%                     & Median & 75\% &  25\%                          & Median & 75\%   &                                   &                                           &                                 &                  &    \\
Simulated Data            & CNN   & 1.20                        & 2.67      & 4.70    & 7.37                              & 16.01      & 27.76      & 2.67                                 & 16.01                                         & 3.22                               & 4.07                & 0.20  \\
                      & U-Net  & \textbf{1.13}                        & \textbf{2.35}      & \textbf{3.99}    & \textbf{7.08}                              & \textbf{14.14}      & \textbf{23.98}      & \textbf{2.35}                                  & \textbf{14.14}                                         & \textbf{2.89}                               & \textbf{3.74}                & \textbf{0.42}  \\
Satellite Data & CNN   & 0.53                     & 0.95   & 1.33 & 33.11                          & 78.22  & 154.05 & 0.98                              & 74.35                                     & 1.57                            & 3.72                & 0.12  \\
                      & U-Net  & \textbf{0.26}                     & \textbf{0.57}   & \textbf{1.01} & \textbf{21.36}                          & \textbf{46.74}  & \textbf{89.40}  & \textbf{0.57}                             & \textbf{46.74}                                    & \textbf{1.22}                            & \textbf{2.47}                & \textbf{0.22}  \\
Combined Data             & CNN   & 0.93                     & 1.50   & \textbf{2.54} & 18.21                          & 46.26  & 148.14 & 1.50                              & 46.26                                     & 2.33                            & 3.41                & 0.82  \\
                      & U-Net  & \textbf{0.48}                     & \textbf{1.07}   & 2.79 & \textbf{14.92}                          & \textbf{37.68}  & \textbf{81.44}  & \textbf{1.07}                             & \textbf{37.68}                                     & \textbf{2.08}                            & \textbf{3.19}                & \textbf{0.86}  
\end{tblr}
\label{tab:result_tab}
\vspace{-10pt}
\end{table}

\textbf{Preliminary Results: }
Table 1 summarizes the performance of the CNN and U-Net models for three dataset scenarios: (a) simulated data, (b) satellite observations, and (c) combined data. The best performance metrics for each scenario are highlighted in bold. The results demonstrate that the proposed U-Net model consistently outperforms the baseline CNN model across all datasets. For the simulated data, the U-Net model achieves significant improvements in MAE, RMSE, and R$^2$ compared to the CNN model. Specifically, the U-Net model achieved a 10\%, 8\%, and 110\% improvement over the CNN in MAE, RMSE, and R$^2$ values, respectively. When applied to satellite observation data, the U-Net model continues to exhibit superior performance in terms of MAE and RMSE, 1.22 and 2.47 respectively, although the R$^2$ values are lower than those obtained with simulated data. This suggests that the complexity of real-world satellite imagery introduces additional challenges for both models. Finally, the combined dataset, incorporating both simulated and satellite data, leads to a substantial improvement in performance, particularly in terms of R$^2$ values. The U-Net model achieves a 330\% improvement over the original benchmark value in R$^2$ when compared to the best result obtained with individual datasets. Both CNN and U-Net models demonstrate strong performance on the combined dataset, but the U-Net model consistently outperforms the CNN model across all metrics.

\vspace{-10pt}
\section{Summary and next steps}
\vspace{-10pt}
We have presented a comprehensive comparison of existing and novel data-driven approaches for quantifying CO$_2$ emission rates using simulated, satellite, and combined datasets. Our work includes the development of a satellite observation-based dataset, its integration with simulated data, and the introduction of a customized U-Net regression model specifically designed for power plant emission rate estimation. The results demonstrate that the U-Net model significantly outperforms the baseline CNN model \cite{le2024deep} across all scenarios, underscoring the effectiveness of our proposed approach. While this is ongoing research, our early findings contribute to advancing data-driven emission rate estimation. Future research directions include: (a) Geographic Expansion: Extending the methodology to additional geographic regions. (b) Plume Quantification: Quantifying CO$_2$ plumes before emission rate estimation and incorporating them as additional input features. (c) Data Fusion: Leveraging high-resolution hyper-spectral satellite data alongside coarse-resolution atmospheric observations to enhance model performance.

\small

\section{Appendix}
\subsection{Data Description}
This paper utilizes simulated power plant emissions data generated by the COSMO-GHG model as part of the SMARTCARB project to replicate CO$_2$ and NO$_2$ retrievals via the Copernicus CO2 Monitoring (CO2M) satellite \cite{le2024deep}. SMARTCARB conducted Observing System Simulation Experiments (OSSEs), which involved high-resolution simulations of CO$_2$, CO, and NO$_2$ for 2015 over Berlin and nearby power plants. These simulations separated CO$_2$, CO, and NO$_2$ into 50 tracers representing different emissions, natural fluxes, and background concentrations, enabling the isolation of plumes from specific sources. Additionally, SMARTCARB developed high-resolution datasets for anthropogenic emissions and biospheric CO$_2$ fluxes, incorporating realistic temporal and vertical variability, including power plant plume rise. A key contribution of this work is the enhancement of the dataset beyond simulated data. Satellite observations from NASA's OCO2/3 missions, which measure column-averaged dry-air carbon dioxide (XCO$_2$) with a 16-day revisit cycle, were processed into gridded daily data for 71 power plant locations. In addition, Sentinel-5P NO$_2$ observations were processed and augmented to align with the features of the simulated data. Weather data, specifically wind conditions, was obtained from TWC. In this study, "satellite data" refers to the combination of XCO$_2$ and NO$_2$ satellite observations with TWC weather data. The simulated and satellite datasets were merged to create a "combined dataset," which was subsequently split into training and testing sets, as explained in Section 3.

\begin{figure}
        \centering
        \includegraphics[width=1.0\linewidth]{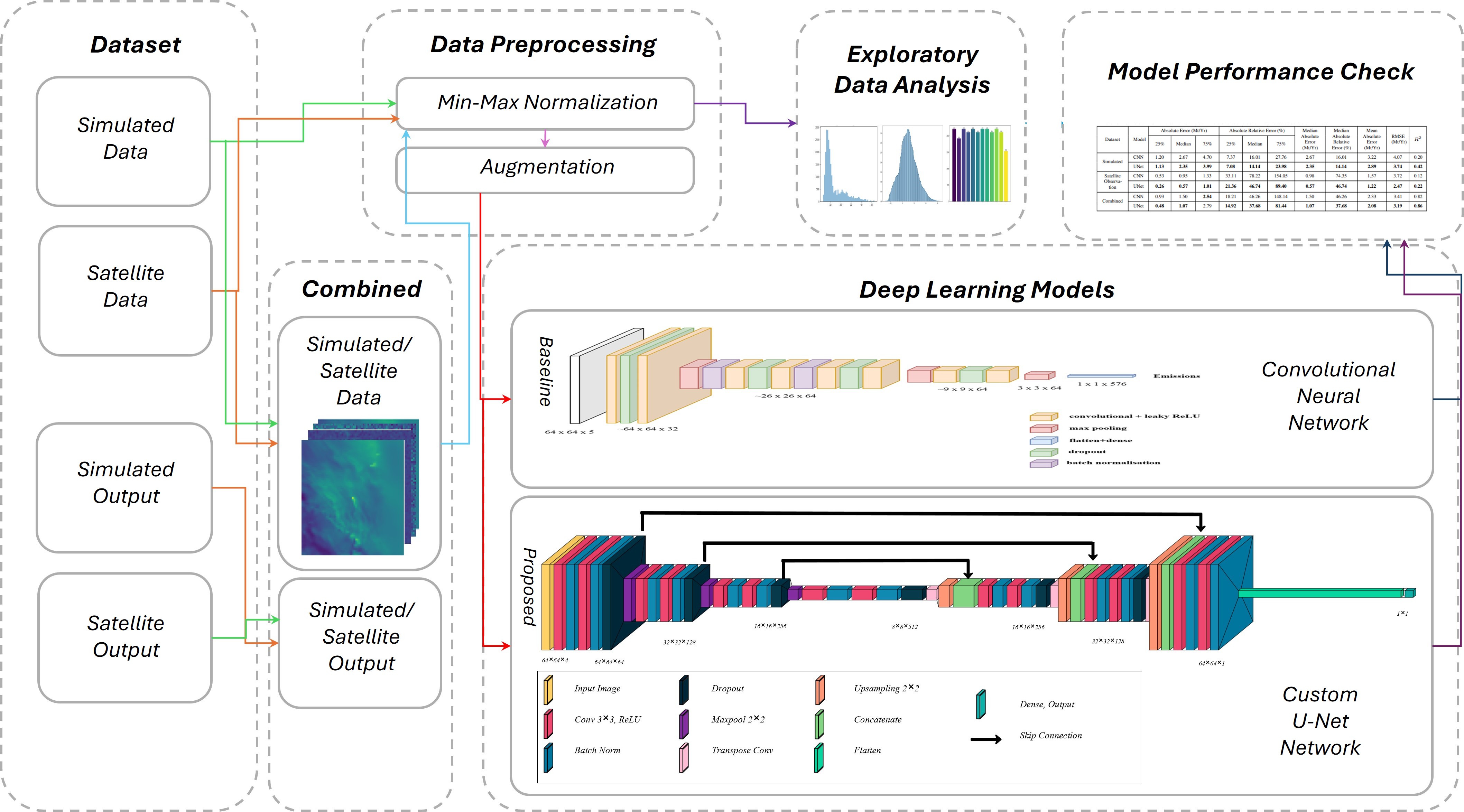}
        \caption{Overview of the proposed architecture}
        \label{fig:overview}
\end{figure}

\subsection{Methodology}
The proposed workflow illustrated in Figure \ref{fig:overview} begins with the analysis of simulated data (case-1) and satellite data (case-2). These datasets undergo pre-processing which includes min-max normalization to standardize the data for subsequent analysis. The normalized outputs are considered as inputs for Exploratory Data Analysis (EDA) where we have tried to uncover patterns, anomalies and correlations. In parallel, the data from Case-1 and Case-2 are merged to form a collective dataset (Case-3) which is pre-processed for EDA. Beyond EDA, the data from Case-1, Case-2 and Case-3 enter the augmentation submodule where techniques like rotation, flipping and scaling are used to enhance the generalizability of the model. The augmented output serves as input to two deep learning models that we are using in our work: (1) the baseline model, which is a conventional CNN architecture adopted from \cite{le2024deep} and (2) the proposed model, which is a custom U-Net network. The summary of these models has been explained in Section 2. Further details are as follows:
In traditional CNN architecture, the process involves progressive downscaling of the image through a series of convolution and pooling to achieve the regression value. Although this technique leads to less computation complexity, it ultimately leads to a loss of spatial information. To overcome the problem of spatial information loss, we utilize the U-Net architecture which addresses the limitation by integrating the skip connections with its encoder-decoder structure. The encoder-decoder architecture can extract features at different scales which also uses its unique skip connection layers, which connect corresponding layers from the encoder to the decoder. The skip connections concatenate feature maps from different layers of the encoder with those from the decoder, effectively retaining spatial information that might otherwise be lost during the down-sampling process. In this way, it ensures the retention of spatial information enabling the model to make better predictions. This structured approach is used to utilize diverse datasets to assess the performance of the model in different cases and demonstrate that the results obtained by our proposed model significantly outperform the conventional CNN baseline model. This is consistent with theory and practical expectations, as the U-Net architecture is more adept at dense prediction tasks. 

\subsection{Code Availability}
The code utilized for the analyzes mentioned in this paper is publicly available on GitHub. To access the repository, visit the link (\href{https://github.com/Dibyabha/dl-co2-pp}{GitHub Link}). The repository includes detailed descriptions and instructions for replicating the results presented in the paper.

\bibliographystyle{plainnat}

\end{document}